\documentclass{article}

    \usepackage[preprint,nonatbib]{neurips_2019}

\usepackage[utf8]{inputenc} 
\usepackage[T1]{fontenc}    
\usepackage{hyperref}       
\usepackage{url}            
\usepackage{booktabs}       
\usepackage{amsfonts}       
\usepackage{nicefrac}       
\usepackage{microtype}      
\usepackage{amsmath}
\usepackage{subfig}
\usepackage{graphicx}
\usepackage{xspace}
\usepackage{booktabs}
\usepackage{tabularx}
\usepackage{multirow}
\usepackage{threeparttable}
\usepackage{color,soul}
\usepackage{bm}
\usepackage{amssymb}

\newcommand{\DeepLABNet}{\textup{\textrm{DeepLABNet}}\xspace}

\title{DeepLABNet: End-to-end Learning of Deep Radial Basis Networks with Fully Learnable Basis Functions}

\author{Andrew Hryniowski$^{1,2,3}$ and Alexander Wong$^{1,2,3}$\\
			$^{1}$ Vision and Image Processing Research Group, University of Waterloo\\
			$^{2}$ Waterloo Artificial Intelligence Institute\\
			$^{3}$ DarwinAI Corp.\\ 			
			\texttt{$\{$apphryni, a28wong$\}$@uwaterloo.ca}
	}

\begin{document}
\maketitle

\vspace{-0.10in}
\begin{abstract}

From fully connected neural networks to convolutional neural networks, the learned parameters within a neural network have been primarily relegated to the linear parameters (e.g., convolutional filters). The non-linear functions (e.g., activation functions) have largely remained, with few exceptions in recent years, \textit{parameter-less}, static throughout training, and seen limited variation in design. Largely ignored by the deep learning community, radial basis function (RBF) networks provide an interesting mechanism for learning more complex non-linear activation functions in addition to the linear parameters in a network. However, the interest in RBF networks has waned over time due to the difficulty of integrating RBFs into more complex deep neural network architectures in a tractable and stable manner. In this work, we present a novel approach that enables end-to-end learning of deep RBF networks with fully learnable activation basis functions in an automatic and tractable manner. We demonstrate that our approach for enabling the use of learnable activation basis functions in deep neural networks, which we will refer to as \DeepLABNet, is an effective tool for automated activation function learning within complex network architectures.

\end{abstract}
\vspace{-0.10in}
\section{Introduction}
\label{sec:label}
\vspace{-0.05in}
The field of neural networks has seen tremendous innovation in the past decade, leading to significant improvements in the scalability as well as representation capabilities of deep neural networks. Such improvements are, in part, due to advanced network architecture designs and improved learning methodologies~\cite{szegedy2016rethinking, he2016deep, ioffe2015batch, kingma2014adam, bengio2009curriculum}. A key insight that improved the representational capabilities of neural networks was successfully stacking simple networks thereby allowing them to more easily learn hierarchical feature representations, leading to the rise of deep learning~\cite{lecun2015deep}. Despite these advances, the classic perceptron~\cite{minsky2017perceptrons}, along with static non-linear functions~\cite{nair2010rectified}, remains at the heart of such networks and are used as the basis of each individual layer design. Even in the most advanced network designs, with a few exceptions~\cite{ramachandran2018searching, jin2016deep, agostinelli2014learning, scardapane2017learning}, individual activation functions remain static from an architecture's inception to its deployment in the field. Largely ignored by the deep learning community in recent years, radial basis function (RBF) networks~\cite{broomhead1988radial} allow unique activation functions for each output, thus providing an interesting mechanism for learning more complex non-linear activation functions. Limited research efforts have investigated integrating RBF networks into a deep learning paradigm. However, such efforts have only utilized RBFs as either a network's output, or one sub-layer in a larger network design~\cite{zadeh2018deep, chen2018deep}; there is no concept of deep hierarchical RBFs.

In this paper we introduce \DeepLABNet, a novel approach that facilitates end-to-end learning of fully trainable deep RBF networks with fully learnable activation functions. Our proposed approach aims to unlocks the power of RBF networks in a hierarchical environment by i) decoupling inter-channel dependencies within each sub-RBF network, ii) using polyharmonic radial basis functions, and iii) strategic regularization and initialization for improved stability during training.

\vspace{-0.1in}
\section{Background}
\label{sec:background}
\vspace{-0.05in}
Network architecture design is a complex process involving many, often iterative, design steps. While certain areas in network design are advancing with rapid progress, such as network macro-architecture (i.e., how many stages are in a network and how they are connected) design~\cite{hu2018squeeze, lin2017refinenet}, other areas have received limited attention, such as non-linear function design~\cite{agostinelli2014learning, ramachandran2018searching}. Throughout this section we review both automated network design and Radial Basis Networks~\cite{broomhead1988radial}.

\vspace{-0.05in}
\subsection{Automated Network Design Methodologies}
\label{sec:learned_actv}
\vspace{-0.05in}
Designing a deep neural network (DNN) can be a time consuming process due to the number of possible configurations and the time required to train each configuration. One must not only decide on the type of architecture (e.g., convolutional neural networks (CNNs)~\cite{lecun1998gradient}, recurrent neural networks (RNNs)~\cite{hochreiter1997long}, etc.), but the detailed architecture design (e.g., number of layers, the type of layers to use, number of neurons per layer, type of activation function to use, etc.). To tackle the bottleneck in network architecture design, recent research efforts have focused on automating the search for optimal network architectures~\cite{zoph2016neural, xie2017genetic, Shafiee2018, stanley2002evolving,wong2018}.

Unlike automated architecture design, automated neuron design methodologies have not received the same amount of research focus. A recent approach to neuron design is to use an intelligent brute force search scheme in which designs are iteratively tested and improved upon~\cite{ramachandran2018searching}. Such a technique is computationally expensive and not tractable for many practitioners. Other research in this area has focused on extending classic neuron designs by parameterizing the neurons with scaling parameters~\cite{he2015delving}, or by using ensemble methods~\cite{harmon2017activation, sutfeld2018adaptive}. These methods are either static or restricted to limited change. To expand the scope of possible learnable neuron designs, piece-wise functions have also been explored~\cite{jin2016deep, agostinelli2014learning, scardapane2017learning, douzette2017b}. A common theme to the learnable activation functions is that they combine peicewise coefficients with a combination of non-linear elements.
\vspace{-0.05in}
\subsection{Radial Basis Networks}
\vspace{-0.05in}
\label{sec:rbf}
The task of supervised learning can be interpreted as an interpolation problem, where one is interpolating between a high-dimensional feature space to a set of class probabilities. A class of neural networks called an Radial Basis Function Networks~\cite{broomhead1988radial} combines RBF interpolation strategies with neural network learning methodologies. Traditionally RBF networks have three layers, an input layer, a single hidden layer, and an output layer.  Given an input vector $\vec{x} \in \mathbb{R}^m$ an RBF network with $h$ neurons will produce a corresponding output vector $\vec{y} \in \mathbb{R}^n$. Within the RBF network the input $\vec{x}$'s distance is measured from the centroid of each hidden neuron. The distance from the $p^{th}$ hidden neuron is used as an input to each basis function
\begin{equation}
\label{eq:rbf_hidden}	
	h_p = \phi(||\vec{x} - \vec{c}_p||)
\end{equation}
where $\phi(\cdot)$ is some kernel function (typically Gaussian), and $\vec{c}_p$ is the kernel's $m$ dimensional centroid. The output $\vec{y}$ of an RBF network is the linear combination of the hidden layer outputs
\begin{equation}
\label{eq:rbf_output}	
	\vec{y} = \bm{\lambda} \vec{h}
\end{equation}
where $\bm{\lambda} \in \mathbb{R}^{n \times p}$, and $\vec{h} \in \mathbb{R}^p$. The \textit{tunable} parameters for an RBF network include the kernel centroids $\{\vec{c}_p\}$ and kernel coefficients $\bm{\lambda}$. In the original RBF paper~\cite{broomhead1988radial} the kernels centroids are either uniformly distributed throughout the input space, set to a subset of the input sample data, or prior knowledge is used to initialize them. By fixing the kernel centroids $\{\vec{c}_p\}$  the kernel coefficients $\bm{\lambda}$ can be learned through over-constrained least squared optimization (assuming a sufficient number of training examples). To provide more flexibility in the set of functions an RBF network can learn, it is common place to add a constant bias $b_n$ for each scalar output.

 \vspace{-0.07in}
\section{Method}
\label{sec:method}
 \vspace{-0.07in}

RBF networks provide a foundation for enabling a neural network to learn more complex, specialized activation functions. However, one of the key issues with traditional RBF networks is that they can be unstable during learning and require too many parameters to be directly utilized in a deep learning context. Traditional RBF networks have three main sets of parameters: i) basis function parameters (e.g., bandwidth of Gaussian basis functions, or the degree of spline basis functions), ii) the radial distance parameters, and iii) the kernel coefficients. Normally, the basis function parameters are predetermined, and the radial distance parameters are initialized either uniformly or to a sub-set of training samples. The kernel coefficients are then learned through linear optimization. These constraints make it extremely difficult to use RBF networks in a hierarchical context (i.e., deep RBF networks).  To mitigate these challenges, \DeepLABNet introduces three key improvements to enable end-to-end learning of deep RBF networks in an automatic (via back-propagation) and tractable manner (scales with model size).

\vspace{-0.05in}
\subsection{Decoupling Inter-Channel Dependencies}
\vspace{-0.05in}
RBF networks utilize \textit{global} basis functions where each basis function is dependent on every input feature and affects every output feature. The multi-input feature dependency of each basis function requires that each basis function use a multi-dimensional centroid. High-dimensional input features increase the computational cost required to find optimal basis centroids and increase the likelihood that the optimization process will get stuck in a local minimum. Furthermore, high-dimensional input features increase the difficulty of initialization of RBF networks. Without a good initialization, a fully learnable RBF network (i.e., when a RBF's centroid is learned with all other parameters) will have difficulty learning and may be unstable, particularly for the case of deep RBF networks.  To tackle this issue, we decouple the input features within an RBF, thus effectively decoupling inter-channel dependencies. Such a decoupling presents three advantages. First, the number of parameters required per layer is significantly reduced (the more channels within a layer, the greater the reduction), thus allowing for reduced computational cost during training and inference, and reduced storage space requirements. Second, by removing inter-channel dependencies, the basis distance functions become scalar, thus allowing for better initialization strategies to be utilized and increased model stability during training. Finally, the reduced computation cost allows additional basis functions to be utilized within a layer allowing for more complex functions to be learned on a per-layer basis while the number of basis functions per channel is reduced (since channels no longer share basis functions). More specifically, \DeepLABNet's feature-level RBF networks leverage the following design:
\vspace{-0.05in}
\begin{equation}
\label{eq:interp_equation}
   y = f(x) = \sum_{i=1}^{s} \lambda_{i} \phi(\vert x - c_{i} \vert) + v_{0} x + v_{1}
\end{equation}
where $x$ and $y$ are the input and output of any given feature-level RBF network, respectively. $\phi(\cdot)$ is the radial basis function, and $c_{i}$ is the $x$ coordinate of the $i^{th}$ control point. $v_{0}$ and $v_{1}$ are scalar components added to the RBF network. The key differences between the proposed definition of a RBF network within \DeepLABNet and that of the original RBF network design in~\cite{broomhead1988radial} are that there are no connections between features, additional per-feature basis functions are added, and a linear component is added via $v_{0}$. Our definition may be seen as having a dedicated RBF network for every feature within the network. For every \DeepLABNet, there are two considerations to be taken into account: i) the number of control points $s$, and ii) the basis function $\phi$ used.

\vspace{-0.05in}
\subsection{Polyharmonic Spline Radial Basis Function}
\vspace{-0.05in}
Selecting which basis function to use is a non-trivial task and decisions regarding such matters often require subject domain expertise. Consider two applications where RBF interpolation is often used, i) approximating a probability distribution~\cite{parzen1962estimation}, and ii) learning a non-linear mapping between 3D coordinate spaces~\cite{szeliski2010computer}. For learning a probability distribution it is common to place a Gaussian kernel at each known sample location. For learning a mapping between coordinate spaces a sparse grid of thin plate spline kernels is generally used~\cite{szeliski2010computer}. A common practice when using RBF networks is to use Gaussian basis functions~\cite{broomhead1988radial}. However, such networks tend to fail when input samples are too distant from a basis function centroid as all basis function output's approach zero. Another class of basis functions are the polyharmonic spline functions, which can be defined as
\begin{equation}
\label{eq:spline_basis_function}	
	U_k(r) =
	\begin{cases}
	r^k, & \text{if $k$ is odd} \\
	r^k \log{r}, & \text{if $k$ is even}
	\end{cases}
\end{equation}
where $r$ is the output of some radial distance function (e.g., Euclidean), and $k$ is the degree of the kernel. Note for all $k$'s that $U_k(0) = 0$. Unlike the Gaussian basis function, polyharmonic basis functions implicitly handle outliers as inputs further from a basis centroid are given additional weighting. Polyharmonic spline functions have not gained traction in the deep learning community as their outputs are not bounded, their outputs grow at a polynomial rate, and these functions can produce unstable networks between updates. Within RBFs, such issues can be mitigated as multiple basis functions, when used in unison, can counteract each other in their respective \textit{unstable} regions. The robustness of the polyharmonic spline basis functions is a crucial aspect of \DeepLABNet that allows RBF networks to be leveraged in a deep hierarchical context.

\vspace{-0.05in}
\subsection{Strategic Initialization and Regularization}
\vspace{-0.05in}
Initialization of any machine learning model can have a large impact on the performance of the model; RBF networks are no exception. Random initialization of an RBF network can result in a function whose gradients quickly explode (depending on the kernel), especially when the basis centroids are positioned near one another, or set everything to zero. To avoid such issues, we utilize a initialization scheme where control point pairs $\{(x, y)_{i \in s}\}$  are uniformly placed along a 'hockey stick' curve and the parameters $\{\lambda_{i \in s}\}$, $v_{0}$, and $v_{1}$ are solved via the following linear system:
\begin{equation}
\label{eq:spline_matrix}	
\begin{bmatrix}
U(\vert \vec{x} - \vec{x}^T \vert)  & \textbf{C} \\
\textbf{C}^T & \textbf{0}
\end{bmatrix}
\cdot
\begin{bmatrix}
\vec{\lambda} \\
\vec{v}
\end{bmatrix}
=
\begin{bmatrix}
\vec{y} \\
\vec{0}
\end{bmatrix}
\end{equation}
where  $\textbf{C} = \begin{bmatrix} \vec{x} & \textbf{1} \end{bmatrix}$, $\vec{x} = [x_{1} \dotsc x_{s}]^T$, $\vec{y} = [y_{1} \dotsc y_{s}]^T$, $\vec{\lambda} = [\lambda_1 \dotsc \lambda_s]^T$, and $\vec{v_a} = [v_{0}\,v_{1}]^T$.

To ensure stability when training the kernel coefficients, $\{\lambda_{i \in s}\}$ must remain balanced, otherwise the RBF output can become overly sensitive to updates to $\lambda_{i}$'s during learning and result in drastically different non-linear functions. Considering Equation~\ref{eq:spline_matrix} one can see three strict constraints

\vspace{-0.05in}
\begin{table}[h]
    \centering
    \begin{tabular}{c c c c c c}
    	1. & $f(x)=y$  &
    	2. & $\sum_{i=1}^{s} \lambda_i x = 0$ &
    	3. & $\sum_{i=1}^{s} \lambda_i = 0$
    \end{tabular}
\end{table}
Inspired by the third constraint, we propose adding a regularization to a network's loss function that minimizes the absolute sum of each RBF feature coefficients in order to help regulate this instability. The loss function is defined as,
\vspace{-0.15in}
\begin{equation}
\label{eq:rbf_reg}
    L_{\DeepLABNet} = L_{model}  + \lambda_{sum}\frac{1}{A} \sum_a^A | \sum_i^{s} \lambda_{ai} |
\end{equation}
where $L_{\DeepLABNet}$ is the loss function for a network using \DeepLABNet, $L_{model}$ is the loss function for a given model without the added RBF functions, and $A$ is the number of RBF functions in a DNN.

\vspace{-0.07in}
\section{\DeepLABNet Hyperparameter Comparison}
\label{sec:exp_results}
\vspace{-0.07in}
\DeepLABNet extends deep learning model architectures by introducing learnable non-linear components via limited (features are not connected), weight-sharing (the non-linearities are convolved across channels), radial basis function networks. The addition of these RBF networks introduces many new design decisions which a practitioner must take into consideration when building a deep neural network, including: the type(s) of kernel(s) used, the number of basis functions per RBF network, how to initialize the RBF parameters, and whether the RBF parameters should be shared between the RBF networks. Throughout this section we explore these considerations by performing a series of comparative analysis' in which a single design decision is varied in each instance.
\vspace{-0.05in}
\subsection{Experimental Setup}
\vspace{-0.05in}
Each variation in \DeepLABNet's design is tested on three architectures: LeNet-5~\cite{lecun1998gradient} is used for MNIST, and ResNet-20~\cite{he2016deep} is used for both CIFAR-10 and CIFAR-100. For the remainder of this paper, unless otherwise stated, the experimental setup for each model is shown in Table~\ref{tab:exp_setup}. Note the right most column 'Weight Sharing' indicates the scope to which the RBF parameters within each \DeepLABNet model are shared, and that L.R. in the first parameter column stands for learning rate.

\begin{table}[h]
     \vspace{-0.15in}
    \caption{\DeepLABNet model experimental setup}
    \centering
    \begin{threeparttable}
    \begin{tabular} { l | c c c c c }
        \toprule
        ~&
        \multirow{1}{*}{\centering\textbf{Initial L.R.}} &
        \multirow{1}{*}{\centering\textbf{$\#$ Epochs}} &
        \multirow{1}{*}{\centering\textbf{$L^2$ rate}} &
        \multirow{1}{*}{\centering\textbf{$\lambda_{sum}$}} &
        \multirow{1}{*}{\centering\textbf{Weight Sharing}}\\
        \midrule
        \multirow{1}{*}{\bf LeNet-5 MNIST}
        & $10^{-3}$ & 30 & $10^{-4}$ & $10^{-2}$ & Layer-wise \\
        \multirow{1}{*}{\bf ResNet-20 CIFAR-10}
        & $10^{-3}$ & 200 & $10^{-4}$ & $10^{-2}$ & Channel-wise \\
        \multirow{1}{*}{\bf ResNet-20 CIFAR-100}
        & $3\times10^{-3}$ & 200 & $3\times10^{-4}$ & $10^{0}$ & Network-wise \\
        \bottomrule
    \end{tabular}
    \end{threeparttable}
    \label{tab:exp_setup}
\end{table}

The remaining setup of each configuration not listed in Table~\ref{tab:exp_setup} is as follows. For the LeNet-5 case the learning rate is reduced by an order or magnitude at 20 epochs and 25 epochs. For the ResNet-20 cases, the learning rate is reduced to $10^{-3}$, $3*10^{-4}$, $10^{-4}$, and $10^{-5}$ at 100, 130, 150, and 175 epochs, respectively. Note that weight decay is not applied to the RBF parameters. Each RBF network within \DeepLABNet uses three basis functions and a polyharmonic spline of degree three (i.e., $r^3$). Each RBF is initialized to a 'hockey-stick' like shape, defined in Section~\ref{sec:init}. Finally, the output of all RBF networks are clipped at $\pm 15$ to help ensure stability during early training.

The Adam optimizer~\cite{kingma2014adam} is used for backpropagation, with random weight initialization for all non-RBF parameters. The training images in each dataset are padded by 4 pixels on all sizes, each image is randomly cropped to the original input dimensions, then randomly flipped along the vertical axis, and finally normalized. A batch size of 128 is used for each network. The standard bias term after convolution, and the Batchnorm~\cite{ioffe2015batch} scale and offset parameters are removed for \DeepLABNet when used on CIFAR-10 and CIFAR-100 as such parameters are redundant.

\vspace{-0.05in}
\subsection{RBF Kernel Comparison}
\vspace{-0.05in}

\begin{figure*}[!t]
    \vspace{-0.2in}
    \centering
    \begin{tabular}{c c c}
        \subfloat[Gaussian $e^{-x^2}$\label{subfig:rbf_guassian}]{\includegraphics[width=0.24\linewidth]{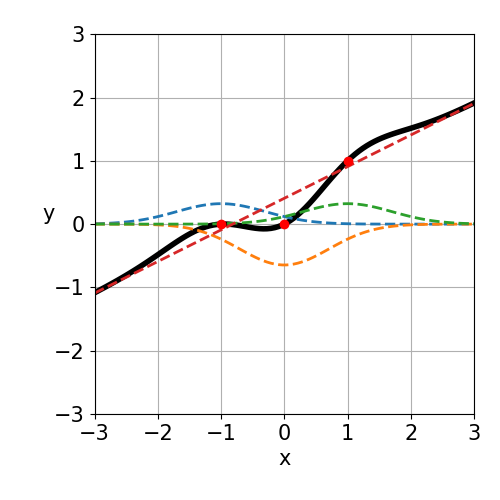}}
        & \subfloat[Multiquadric $\sqrt{1 + x^2}$ \label{subfig:rbf_multiquad}]{\includegraphics[width=0.24\linewidth]{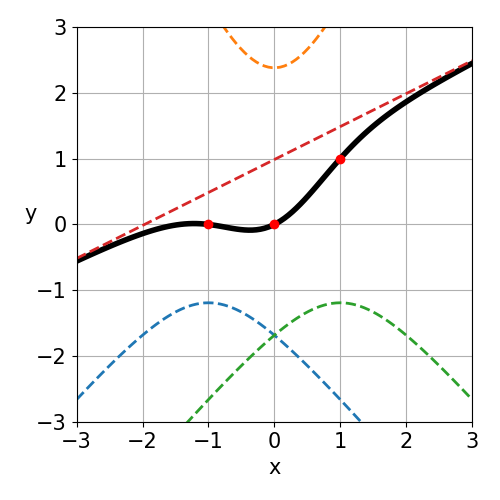}}
        & \subfloat[Spline $x^3$
        \label{subfig:rbf_spline}]{\includegraphics[width=0.24\linewidth]{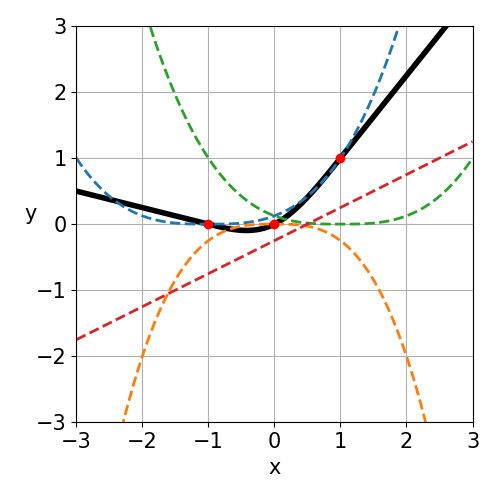}}
    \end{tabular}

    \caption{\small Three different RBF networks used with \DeepLABNet at initialization. Each function is initialized to the hockey-stick shape and is shown in black. The components of the RBF networks are displayed in coloured dotted lines. The non-linear components of each RBF network are coloured in blue, green, and orange, and the linear component of each RBF network is coloured in red. The red dots are the location of each kernel's centroid.}
    \vspace{-0.2in}
    \label{fig:rbf_kernels}
\end{figure*}

The primary building block of an RBF network is its kernel. The kernel has a larger effect on how well an unknown function can be approximated and the speed of convergence. We compare three variants of \DeepLABNet, each one using a different type of RBF kernel. The three kernels are a Gaussian kernel, a multiquadric kernel, and a polyharmonic spline kernel of degree 3. A visualization of each kernel is shown in Figure~\ref{fig:rbf_kernels}. The results of the comparison are shown in the table below.

\begin{table}[h]
    \vspace{-0.1in}
    \caption{Comparison of the different types of kernels used in \DeepLABNet (Accuracy (\%))}
    \centering
    \begin{threeparttable}
    \begin{tabular} { l | c c c }
        \toprule
        ~&
        \multicolumn{1}{p{2.0cm}}{\centering\textbf{Gaussian}} &
        \multicolumn{1}{p{2.0cm}}{\centering\textbf{Multiquadric}} &
        \multicolumn{1}{p{2.0cm}}{\centering\textbf{Spline}} \\
        \midrule
        \multirow{1}{*}{\bf LeNet-5 MNIST}
        & 98.7 & 98.7 & \textbf{98.9}\\
        \multirow{1}{*}{\bf ResNet-20 CIFAR-10}
        & 90.8 & 91.3 & \textbf{91.5} \\
        \multirow{1}{*}{\bf ResNet-20 CIFAR-100}
        & 65.3 & 65.4 & \textbf{66.3}\\
        \bottomrule
    \end{tabular}
    \vspace{-0.1in}
    \end{threeparttable}
    \label{tab:kernel_comparison}
\end{table}
In each case the models that use the polyharmonic spline kernel demonstrated superior results and the models that use the  Gaussian kernel performed the worst (or tied for worst). For the MNIST model all three kernels where within $0.2\%$ error of each other. For the CIFAR-10 model the Gaussian kernel under performed the other kernels by at least $0.5\%$. With the CIFAR-100 model the polyharmonic spline kernel had a performance gap of $0.9\%$ compared to the next closest kernel. In addition to their lower performance, the Gaussian kernel models also demonstrated a slower convergence rate compared to the other kernels. Such behavior is likely due the limited scope of each individual kernel in that the response of each kernel approaches 0 as the input moves away from the kernels centroid.
\vspace{-0.05in}
\subsection{Number of Kernels per RBF}
\vspace{-0.05in}
Theoretically, an RBF network has the ability to approximate more complex functions as the number of kernels $s$ per RBF increases. To test this hypotheses we measured the performance of \DeepLABNet when 3, 5, and 7 kernels per RBF network are used; the same number of kernels per RBF is used within \DeepLABNet at a time. Experimentally, using 3 kernels per RBF network provided the best validation performance. Increasing the number of kernels per RBF only resulted in increased over fitting on the training data while often hurting the validation performance. In addition, adding kernels to the RBF networks significantly increases the time required to train any given model. While adding kernels under the current setup was not fruitful it is still possible that better regularization on the the RBF kernels may help extract additional performance when using more than 3 kernels per RBF network. Other than the regularization of Equation~\ref{eq:rbf_reg} \DeepLABNet has no constraints on the RBF kernel parameters. As such, \DeepLABNet is free to place the kernels anywhere it \textit{learns}. Consider two kernels learning a placement on top of each other (i.e., their centroids are equal). \DeepLABNet is at best performing a redundant calculation, since the kernels could be easily combined without loss of precision, and at worst the two kernels cancel each other. Further investigation of kernel placement density may illuminate more effective strategies for extracting additional performance from high-kernel-count RBF networks within \DeepLABNet.

\vspace{-0.05in}
\subsection{Strategic Initialization}
\label{sec:init}
\vspace{-0.05in}
\DeepLABNet with random weight initializations tend to be numerically unstable during early training. Such instability can be attributed to two kernel centroids being close to one another with large and opposite $\lambda_i$'s. To avoid this issue we uniformly distribute the kernel centroids along the x-axis in the range of $[-r, r]$ and bound the $\{\lambda_i\}'s$ through strategic initialization. We test three different initialization strategies: linear, random-y, and hockey-stick. Linear initialization is the identity operation where $\{\lambda_i\}$ and $v_1$ are set to zero, and is set $v_0$ to one. Random initialization generates a set of control points $\{(x_i, y_i)\}$ by uniformly sampling each $y_i$ from the range $[-r, r]$, and then solving for RBF parameters using Equation~\ref{eq:spline_matrix}. Finally, like random-y, hockey-stick initialization selects a set of control points $\{(x_i, y_i)\}$  and solves the RBF parameters. For hockey-stick the initialization parameters are uniformly spaced long a ReLU curve. The order of performance across all three models is consistent between the initialization strategies. Hockey-stick initialization strategy provides the best performance, followed by random-y, with linear initialization performing the worst. Interestingly, the the linear-initialized RBF networks quickly learn non-linear components early in training, but such initialization still results in worst performance compared to the other initialization methods. Despite \DeepLABNet's ability to automatically learn activation functions, RBF initialization and its effect on early learning play an important part in a model's final performance.

\begin{table}[h]
     \vspace{-0.1in}
    \caption{Comparison of the different weight initializations (Accuracy (\%))}
    \centering
    \begin{threeparttable}
    \begin{tabular} { l | c c c }
        \toprule
        ~&
        \multicolumn{1}{p{2.0cm}}{\centering\textbf{Linear}} &
        \multicolumn{1}{p{2.0cm}}{\centering\textbf{Random-y}} &
        \multicolumn{1}{p{2.5cm}}{\centering\textbf{Hockey-Stick}} \\
        \midrule
        \multirow{1}{*}{\bf LeNet-5 MNIST}
        & 98.6 & 98.6 & \textbf{98.9} \\
        \multirow{1}{*}{\bf ResNet-20 CIFAR-10}
        & 90.5 & 91.0 & \textbf{91.5}   \\
        \multirow{1}{*}{\bf ResNet-20 CIFAR-100}
        & 63.8 & 65.3 & \textbf{66.3}  \\
        \bottomrule
    \end{tabular}
    \end{threeparttable}
    \label{tab:init_searc}
         \vspace{-0.1in}
\end{table}

\vspace{-0.05in}
\subsection{Weight Sharing}
\vspace{-0.05in}

\begin{figure*}[!t]
     \vspace{-0.1in}
	\centering
	\includegraphics[width=0.90\linewidth,trim={3cm 0 3cm 0},clip]{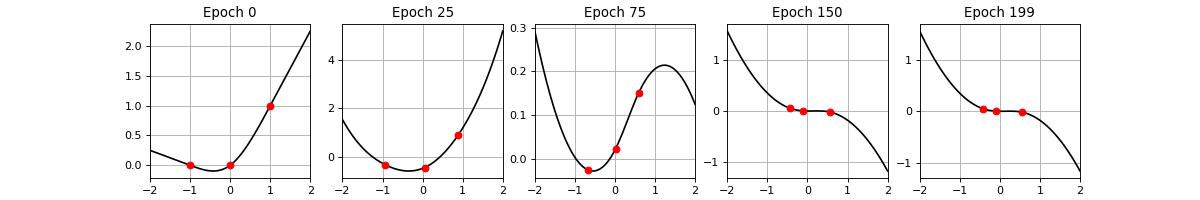}
	\caption{\small Example of a \DeepLABNet RBF activation function throughout training. This RBF function uses the polyharmonic spline kernel with three control points (red dots). The function (initialized based on a 'hockey stick' curve) is slowly deformed throughout training. Note the $y$ component of the control points are calculated using the learned RBF parameters; they are not directly used during training or inference. }\vspace{-0.2in}
	\label{fig:learned_actv}
\end{figure*}

Standard deep neural networks use the same static activation function across the entire network. Currently the standard activation function is ReLU. Implicit in the shared global activation function design is the assumption that there is no need to have specialized activation functions, as a series of such universal approximator functions can approximate any other function. From a parametrization perspective, all activation functions share the same set of implicit weights. For ReLU such weights would encapsulate the description "when the input is negative the out is zero, when the input is positive the output equals the input". From a network architecture perspective there is no strict underlying reason to enforce such a design constraint. For example, one layer in a network could use the ReLU function while another layer in the network could use a leaky-ReLU function~\cite{xu2015empirical}. In fact, each layer or even each feature in a network could have a unique dampening factor.  We explore three weight sharing granularities, including activation parameters unique to each feature/channel, activation parameters shared within a layer, and activation parameters shared across the entire network.

\begin{table}[h]
    \vspace{-0.15in}
    \caption{Comparison of the different shared weight granularities}
    \centering
    \begin{threeparttable}
    \begin{tabular} { l  l | c c c }
        \toprule
        ~&~&
        \multicolumn{1}{p{2.0cm}}{\centering\textbf{Channel}} &
        \multicolumn{1}{p{2.0cm}}{\centering\textbf{Layer}} &
        \multicolumn{1}{p{2.0cm}}{\centering\textbf{Global}} \\
        \midrule
        \multirow{2}{*}{\bf LeNet-5 MNIST}
        & \textbf{\# Params. (K)}
        & 63.5 &  61.7 & 61.7 \\
        & \textbf{Accuracy (\%)}
        & 98.7 & \textbf{98.9} & 98.6 \\
        \midrule
        \multirow{2}{*}{\bf ResNet-20 CIFAR-10}
        & \textbf{\# Params. (K)}
        & 278 &  273 &  273 \\
        & \textbf{Accuracy (\%)}
        & \textbf{91.5} & 91.3 & 91.2  \\
        \midrule
        \multirow{2}{*}{\bf ResNet-20 CIFAR-100}
        & \textbf{\# Params. (K)}
        & 284 & 279 & 279 \\
        & \textbf{Accuracy (\%)}
        & 65.6 & 66.0 & \textbf{66.3}  \\
        \bottomrule
    \end{tabular}
    \end{threeparttable}
    \label{tab:actv_gran_searc}
         \vspace{-0.15in}
\end{table}
The validation error and the number of parameters per configuration is shown in Table~\ref{tab:actv_gran_searc}. As one can see, the optimal activate weight sharing granularity used is dependent on the under lying model architecture. While some networks like LeNet-5 MNIST demonstrate little variation between tests, ResNet-20 CIFAR-100 varies by $0.7\%$ on the top-1$\%$ error. For ResNet-20 CIFAR-100 model it can be observed that having a single global activation provides superior performance while minimizing the number of parameters with the model. When learning the ResNet-20 CIFAR-100 model the global activation function provides approximately $4\%$ improved performance during early training. This performance advantage was not observed for the channel and layer activation granulatities or with the other models. Furthermore, the large performance gap in early training gradually disappears as the learning rate for each model is reduced.
\vspace{-0.07in}
\section{\DeepLABNet vs RBF Networks}
\label{sec:exp_comp}
\vspace{-0.07in}
We continue our analysis by comparing \DeepLABNet to two other RBF network designs: i) a traditional single layer RBF network, and ii) a deep convolutional neural network that uses traditional RBF networks for activation (for brevity, this network is referred to as the RBF activation network). A standard deep convolutional neural network with ReLU activation is used as a baseline reference. Each of the network design is tested against the following three benchmark datasets: MNIST~\cite{lecun1998mnist}, CIFAR-10~\cite{krizhevsky2009learning}, and CIFAR-100~\cite{krizhevsky2009learning}. The design decisions and training parameters for the test networks are as follows. The single layer RBF networks use 250 hidden neurons for the MNIST dataset, and 1000 hidden neurons for both the CIFAR-10 and CIFAR-100 datasets. The hidden neuron centroids are initialized to a class-wise uniform distribution of training samples. The remainder of the RBF parameters are solve for using least-squares optimization. The baseline ReLU networks use a weight decay of $10^{-4}$ and otherwise follows the \DeepLABNet training procedure established in Section~\ref{sec:exp_results}. Table~\ref{tab:experiment_results} shows the mean validation classification error on the 12 model-dataset combinations.

\begin{table}[h]
    \vspace{-0.15in}
    \caption{Comparison of the different types of RBF networks}
    \centering
    \begin{threeparttable}
    \begin{tabular} { l  l | c | c c c }
        \toprule
        ~&~&
        \multicolumn{1}{p{1.2cm}|}{\centering\textbf{Baseline}} &
        \multicolumn{1}{p{1.5cm}}{\centering\textbf{RBF Network}} &
        \multicolumn{1}{p{2.5cm}}{\centering\textbf{RBF Activation Network}} &
        \multicolumn{1}{p{2.0cm}}{\centering\textbf{\DeepLABNet}} \\
        \midrule
        \multirow{3}{*}{\bf MNIST}
        & \textbf{FLOPs (M)}
        & 0.85 & \textbf{0.77} & 1.25 & 0.95 \\
        & \textbf{\# Params. (K)}
        & \textbf{61.7} & 199 & 105 & 61.7 \\
        & \textbf{Accuracy (\%)}
        & 97.9 & 94.3 & 69.7 & \textbf{98.9} \\
        \midrule
        \multirow{3}{*}{\bf CIFAR-10}
        & \textbf{FLOPs (M)}
        & 82.0 & \textbf{9.24} & 107 & 84.8 \\
        & \textbf{\# Params. (K)}
        & \textbf{272} & 3082 & 339 & 278 \\
        & \textbf{Accuracy (\%)}
        & 90.6 & 51.5 & 33.4 & \textbf{91.5} \\
        \midrule
        \multirow{3}{*}{\bf CIFAR-100}
        & \textbf{FLOPs (M)}
        & 82.0 & \textbf{9.42} & 107 & 84.8 \\
        & \textbf{\# Params. (K)}
        & \textbf{278} & 3172 & 345 & 279 \\
        & \textbf{Accuracy (\%)}
        & 65.4 & 22.8 & 3.55 & \textbf{66.3} \\
        \bottomrule
    \end{tabular}

    \end{threeparttable}
    \vspace{-0.1in}
    \label{tab:experiment_results}
\end{table}
It can be observed that \DeepLABNet has the best average performance on all three datasets. In addition, the baseline network has the smallest number of parameters, while the RBF network has the smallest number of FLOPs. \DeepLABNet outperforms the baseline network by $1.0\%$, $0.9\%$, and $0.9\%$ on MNIST, CIFAR-10, and CIFAR-100, respectively. Despite the number of parameters within the RBF network it was unable to reach performance of the base line model and \DeepLABNet. This result is not surprising as the RBF network is single layer. On the other hand, the RBF activation network is as deep as \DeepLABNet while having more parameters but still fails to outperform even the single layer RBF network. The poor performance of the RBF activation network is mostly due to the network getting stuck in a local minimum during training. Moreover, this lack in performance effectively highlights the difficulty of naively integrating RBF networks directly in a deep learning pipeline and demonstrates \DeepLABNet's merits.

\vspace{-0.07in}
\section{\DeepLABNet Scalability}
\label{sec:larger_models}
\vspace{-0.07in}
The final set of tests in this work investigates how \DeepLABNet performs on both smaller and larger ResNet architectures. To that effect \DeepLABNet is compared to the baseline ReLU configuration for ResNet models of size 8, 14, 20, 32, and 50. Both \DeepLABNet and the baseline model follow the same test configurations as in Section~\ref{sec:exp_comp}. The only difference is that the ResNet models of size 8, 14, 20, 32, and 50 are trained for 150, 175, 200, 235, and 280 epochs, respectively. The drops in the respective learning rates are scaled linearly with the number epochs trained. The results for this test are shown in the table below (note \DeepLABNet is abbreviated to DLN).

\begin{table}[h]
\setlength{\tabcolsep}{2.25pt}
    \vspace{-0.1in}
    \caption{Comparison of ResNet models across varying model sizes}
    \centering
    \begin{threeparttable}
    \begin{tabular} { l l | c c | c c | c c | c c | c c}
        \toprule
        \multicolumn{2}{c|}{\multirow{2}{*}{\textbf{ResNet-$\#$}} } &
        \multicolumn{2}{p{1.8cm}|}{\centering\textbf{8}} &
        \multicolumn{2}{p{1.8cm}|}{\centering\textbf{14}} &
        \multicolumn{2}{p{1.8cm}|}{\centering\textbf{20}} &
        \multicolumn{2}{p{1.8cm}|}{\centering\textbf{32$^*$}} &
        \multicolumn{2}{p{1.8cm}}{\centering\textbf{50$^*$}} \\
        ~&~& ReLU & DLN & ReLU & DLN & ReLU & DLN & ReLU & DLN & ReLU & DLN \\
        \midrule
        \multirow{3}{*}{\bf CIFAR-10}
        & \textbf{FLOPs (M)}
        & 25.1 & 26.2 & 53.5 & 55.5 & 81.9 & 84.7 & 139 & 143 & 224 & 231 \\
        & \textbf{\# Params. (K)}
        & 77.7 & 80.0 & 175 & 179 & 272 & 278 & 466 & 477 & 757 & 775 \\
        & \textbf{Accuracy (\%)}
        & 86.6 & 87.5 & 89.7 & 90.7 & 90.6 & 91.5 & 91.5 & 91.6 & 91.6 & 91.9  \\
        \midrule
        \multirow{3}{*}{\bf CIFAR-100}
        & \textbf{FLOPs (M)}
         & 25.1 & 26.2 & 53.5 & 55.5 & 82.0 & 84.8 & 139 & 143 & 224 & 231 \\
        & \textbf{\# Params. (K)}
         & 83.6 & 83.9 & 180 & 181 & 278 & 279 & 472 & 474 & 763 & 766  \\
        & \textbf{Accuracy (\%)}
         & 53.9 & 59.6 & 64.2 & 64.4 & 65.4 & 66.3 & 67.2 & 67.8 & 67.5 & 68.3  \\
        \bottomrule
    \end{tabular}
    \begin{tablenotes}
        \small
        \item $^*$ the results of these models are the average of three separate runs
    \end{tablenotes}
    \end{threeparttable}
    \label{tab:larger_models}
    \vspace{-0.10in}
\end{table}

For each model comparison \DeepLABNet provides superior performance compared to the baseline ReLU model. Most of the performance differences are within $1.0\%$. The smallest performance difference is $0.1\%$ on the ResNet-32 CIFAR-10 model, while the greatest performance difference is $5.7\%$ on the ResNet-8 CIFAR-100 model. The large difference between small models trained on the more difficult dataset indicates that \DeepLABNet may be particularly useful for computing on the edge (e.g., in applications where model size, and inference latency are important considerations). Another interesting trend is that \DeepLABNet provides both ResNet-32 models equal or greater performance compared to the respective ReLU ResNet-50 models, furthering \DeepLABNet's potential utility for maintaining performance while reducing model size. These results clearly demonstrate \DeepLABNet's merits across a variety of model sizes.

The performance gains of \DeepLABNet come with three important considerations. i) \DeepLABNet has a significant effect on the time required to train the above models. With ResNet-8, and ResNet-50 \DeepLABNet cause a $50\%$, and a $150\%$ increase in training time when using Tensorflow~\cite{tensorflow2015-whitepaper} with an Nvidia GTX 1080 Ti. ii) As noted in a previous section, \DeepLABNet is more prone to overfitting than the ReLU based model, and increasingly so the larger a model becomes. Experimentally we observed that the overfitting is correlated with the length of time spent a given learning rate. 3) \DeepLABNet is not as stable as the ReLU based models. While the average performance of \DeepLABNet is superior to the ReLU models, the variance in the performance is also greater, thus potentially requiring a greater amount of training runs. Comparable with overfitting, variance in model performance also increases with model size.

\vspace{-0.05in}
\section{Conclusions}
\label{sec:conclu}
\vspace{-0.05in}

The comparisons throughout this work demonstrate that \DeepLABNet effectively unlocks the potential of RBF networks in a modern deep learning paradigm. \DeepLABNet provides superior performance over both standard deep neural network and single layer RBF networks, and while overcoming the limitations of hierarchical RBF networks. These improvements require limited additional learning parameters and FLOPs, increased but acceptable training times, and additional sensitivity to a models learning regimen compared to ReLU based models. \DeepLABNet achieves its improved performance and flexible activation design by replacing static network components with learnable non-linear components thus allowing for a more encompassing end-to-end learning experience. The performance of \DeepLABNet indicates that there is potential room for improving automated neuron design procedures, thus making it an attractive candidate for future research. As such, \DeepLABNet was shown to be an effective tool for automated activation function learning within complex network architectures.

\medskip
\small

\bibliographystyle{ieeetr}
\bibliography{refs}

\end{document}